\UseRawInputEncoding
\documentclass[letterpaper, 10 pt, journal, twoside]{IEEEtran}

\IEEEoverridecommandlockouts                              

\usepackage{amsmath}
\usepackage{graphicx}
\usepackage{multirow}
\usepackage{booktabs}
\usepackage{gensymb}
\usepackage{braket}
\usepackage{amssymb}
\usepackage{physics}
\usepackage{caption}
\usepackage{subcaption}
\usepackage{algorithm} 
\usepackage{algpseudocode}
\usepackage[hidelinks]{hyperref}
\usepackage[noabbrev,capitalize]{cleveref}
\usepackage{ragged2e}
\usepackage{array}
\usepackage[]{fancyhdr}
\newcolumntype{R}{>{\RaggedLeft\arraybackslash}p{3.5em}}
\newcolumntype{G}{>{\RaggedLeft\arraybackslash}p{3.5em}}
\title{
    Unlocking the Performance of Proximity Sensors\\by Utilizing Transient Histograms
}

\author{Carter Sifferman$^{1}$, Yeping Wang$^{1}$, Mohit Gupta$^{1}$, and Michael Gleicher$^{1}$%
\thanks{Manuscript received: May 31, 2023; Revised August 19, 2023; Accepted August 23, 2023.}
\thanks{This paper was recommended for publication by Editor Tamim Asfour upon evaluation of the Associate Editor and Reviewers' comments.
This work was supported by Los Alamos National Laboratory and the Department of Energy, a University of Wisconsin Vilas Award, and National Science Foundation awards 1830242, 1943139, 2107060, and 2003129.

2023 IEEE. Personal use of this material is permitted.  Permission from IEEE must be obtained for all other uses, in any current or future media, including reprinting/republishing this material for advertising or promotional purposes, creating new collective works, for resale or redistribution to servers or lists, or reuse of any copyrighted component of this work in other works.
}
\thanks{$^{1}$The authors are with the Department of Computer Sciences,
         University of Wisconsin-Madison, Madison 53706, USA
        {\tt\footnotesize [sifferman|yeping|mohitg|gleicher]\@cs.wisc.edu}}%
}

\begin{document}
\maketitle
\begin{abstract}
We provide methods which recover planar scene geometry by utilizing the \textit{transient histograms} captured by a class of close-range time-of-flight (ToF) distance sensor. A transient histogram is a one dimensional temporal waveform which encodes the arrival time of photons incident on the ToF sensor. Typically, a sensor processes the transient histogram using a proprietary algorithm to produce distance estimates, which are commonly used in several robotics applications. Our methods utilize the transient histogram directly to enable recovery of planar geometry more accurately than is possible using only proprietary distance estimates, and consistent recovery of the albedo of the planar surface, which is not possible with proprietary distance estimates alone. This is accomplished via a differentiable rendering pipeline, which simulates the transient imaging process, allowing direct optimization of scene geometry to match observations. To validate our methods, we capture 3,800 measurements of eight planar surfaces from a wide range of viewpoints, and show that our method outperforms the proprietary-distance-estimate baseline by an order of magnitude in most scenarios. We demonstrate a simple robotics application which uses our method to sense the distance to and slope of a planar surface from a sensor mounted on the end effector of a robot arm.
\end{abstract}
\vspace{-2em}

\section{Introduction}

\begin{figure}
    \centering
    \includegraphics[width=\linewidth]{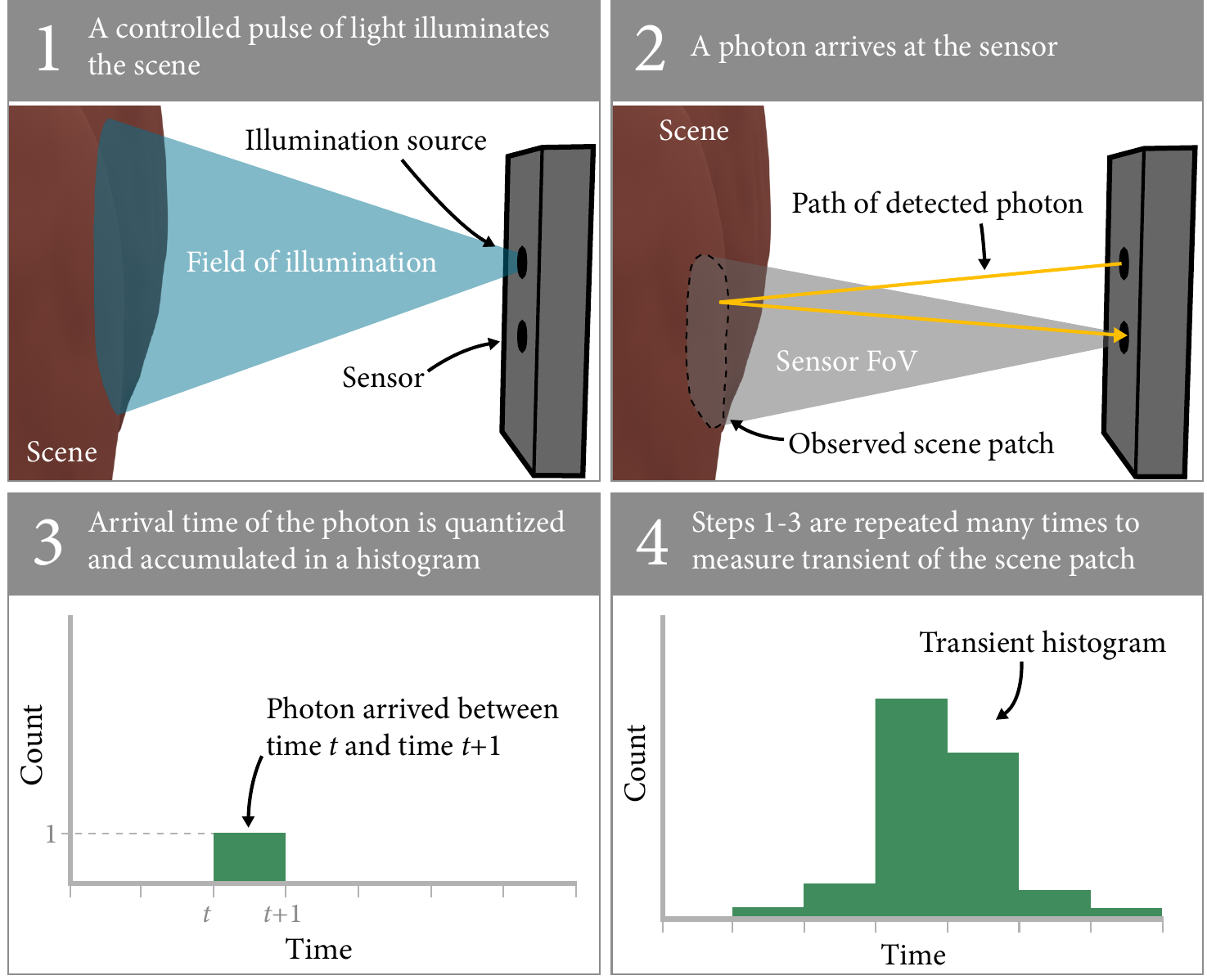}
    \caption{\textbf{The transient histogram is a temporal waveform which records the response of a scene patch when exposed to a pulsed light source.} Presently available commodity sensors estimate the transient histogram through a repeated process.}
    \label{fig:sensor_diagram}
    \vspace{-1.5em}
\end{figure}

\IEEEPARstart{O}{ptical} time-of-flight proximity sensors which measure scene \textit{transients} have recently become widely available. These sensors operate by illuminating the scene with a pulse of light, and measuring the \textit{shape} of that pulse over time as it returns back from the scene in a \textit{transient histogram}, as shown in \cref{fig:sensor_diagram}. These \textit{transient sensors} have seen use in robotics due to their ability to reliably report a distance estimate over a wide range (1cm - 5m) while being small ($<20$ mm$^3$), lightweight, and low-power (on the order of milliwatts per measurement) \cite{TMF8820, VL6180X}. Because of their form factor, transient sensors can be placed in locations where higher resolution 3D sensors cannot, such as on the gripper or links of a robot manipulator, or on very small robots. While these sensors have many desirable properties, existing robotics applications do not utilize the transient histograms, instead relying on low-resolution (at most $4\cross4$ pixel) proximity measurements generated onboard the sensor. Due to the coarseness of their measurements, these sensors are presently only used in robotics for coarse sensing, \textit{e.g.,} detecting the presence of obstacles or distance to a target.

In this work, we utilize transient histograms directly to recover accurate planar scene geometry, and consistent planar albedo from a single $3\cross3$ transient sensor measurement. Planar geometry is an initial use case for our methods, and is a special case of 3D sensing that has many applications in robotics. A robot interacting directly with any planar surface will benefit from sensing the geometry of that surface accurately and at a close range. For example: a robot arm placing an object on a tabletop, sweeping a floor, or writing on a flat surface; a mobile robot localizing the floor and walls of a room; or a drone finding a safe spot to land. Our method enables accurate recovery of this planar geometry that otherwise would have required multiple proximity sensors or a depth camera, while maintaining the same very small form factor and operating at ranges as low as 1cm.

This work is the first to demonstrate that utilizing transient histograms can improve the performance of proximity sensors over utilizing proprietary on-sensor distance estimates. To achieve this, our contributions are 1) an effective \textit{forward imaging model} for commodity proximity sensors, 2) a \textit{differentiable rendering pipeline} which implements the forward imaging model and utilizes it to recover planar geometry and albedo directly from transient histograms, 3) an \textit{empirically calibrated approach} which approximates the performance of the differentiable rendering pipeline and acts as a baseline, and 4) \textit{empirical evidence} that our approaches outperform alternative methods which do not utilize transient histograms.

We present two methods for recovery of planar geometry, one of which can also be used to consistently recover the albedo of the planar surface. To evaluate our methods, we gather thousands of measurements of eight planar surfaces with a commodity transient sensor from a range of angles-of-incidence and distances. We find that our methods which utilize the transient histogram are more accurate and robust than those which rely on proprietary distance estimates. We also find that our method recovers consistent planar albedo, which is not possible to recover from proprietary distance estimates, as they do not encode intensity information. We build a demonstration application which takes advantage of the small size of a transient sensor by mounting the sensor to the gripper of a robot arm. Measurements from the sensor are used to measure the distance to the surface below the gripper and to ensure that the surface is level before placing an object.

\section{Background: Transient Histograms}
A \textit{transient} is a one dimensional temporal waveform which measures the light reflected from a scene over time in response to a pulsed light source. Recently, sensors which are able to capture a transient quantized over short (picosecond) time scales have become available for distance/range measurement using the time-of-flight principle. We refer to these sensors as \textit{transient sensors}. These sensors come in a range of form factors: from high resolution lab-grade arrays to mobile device LiDAR modules, to very small proximity sensors. Most notable of the currently available transient sensors is the single photon avalanche diode (SPAD) \cite{cova1996avalanche, pellegrini2000laser}, which is inexpensive and commonly used in robotics (see \cref{subsec:transient_in_robotics}).

As shown in \cref{fig:sensor_diagram}, a SPAD-based sensor approximates the transient histogram through a repeated process. A controlled pulse of light (typically infrared) flood-illuminates the scene in front of the sensor. Each sensor pixel records the elapsed time between this pulse being sent and a single photon arriving at the pixel. This arrival time is quantized to a discrete bin and accumulated in a \textit{transient histogram}. Over many photon arrivals, this histogram approximates the true transient. In practice, a commodity sensor may record millions of photon arrivals to form a transient histogram. In sensors with an array of pixels, a transient histogram is generated for each pixel.

Currently available commodity transient sensors have many desirable properties. Many are capable of gathering transient histograms at 30 frames per second. Maximum range varies by model, but may be as high as 5m, with a typical minimum range of 1cm. There exist techniques for mitigating the effects of high ambient light on these sensors, enabling their operation in diverse environments \cite{pediredla2018signal, gupta2019asynchronous}.

In this work, we evaluate our method using the SPAD-based TMF8820 sensor manufactured by AMS. We choose this sensor because it 1) allows access to transient histograms through an official driver, 2) captures a $3\times3$ grid of transient histograms at a time, each from a different region of its field-of-view, and 3) provides access to a ``reference histogram'' which encodes the intensity of the laser pulse over time. In the sensor's default configuration, transient histograms are summarized onboard the sensor via a proprietary algorithm, and a distance and confidence estimate are reported for each field-of-view region. We reconfigure the sensor to report a transient histogram \textit{and} proprietary distance estimate for each FoV region. While we utilize the TMF8820 in this work, the methods we propose can be applied to any sensor which reports a transient histogram.

\section{Related Work}
\subsection{Transient Sensors in Robotics}
\label{subsec:transient_in_robotics}
Transient sensors are widely used in robotics applications as they provide highly reliable distance measurements, while being lightweight, low-cost and low-power. Tsuji and Kohama \cite{Tsuji2019} demonstrate a ``sensitive skin'' for a robot arm consisting of many single-pixel transient sensors. Similarly, Adamides et al. \cite{adamides2019time} propose an array of transient sensors mounted around a robot wrist to achieve safe human-robot collaboration. Escobedo et al. place transient sensor on robot joints and  use them to actively avoid collisions \cite{escobedo2021contact}. Transient sensors have been used to detect obstacles when mounted on a drone \cite{tsuji2022omnidirectional}. Our previous work characterized two transient sensors and demonstrated a method for extrinsically calibrating their position relative to a robot arm to which they are attached \cite{sifferman2022}. Outside of robotics, commodity transient sensors have seen use in wearable computing \cite{selvaraj2018stair} and inspection applications \cite{marko2020distance}. In these prior works, only the sensor's proprietary distance estimates are utilized. To our knowledge, our work is the first to utilize the transient histogram in a robotics setting.



\subsection{Inference from Transient Histograms}
Our differentiable rendering pipeline and forward imaging model are heavily inspired by prior work in imaging. Photon arrival times, like those encoded by a transient histogram, are heavily utilized in non-line-of-sight (NLOS) imaging, pioneered by Velten et al. \cite{velten2012recovering}. In NLOS imaging, scene geometry is recovered from around the corner by reflecting a powerful pulsed laser off a diffuse surface. Recent NLOS works utilize the same single photon avalanche diode (SPAD) technology as the sensor that we use in this work \cite{buttafava2015non, shen2021NeTF}. However, the imaging setup used in NLOS imaging requires a high-powered laser and relatively large, expensive laboratory grade SPAD sensors, which have thousands of histogram bins and very precise timing. In contrast, the sensor that we use in this work is readily available, small, lightweight, and eye safe, but reports only 128 histogram bins, and has less precise timing and optical characteristics.

A number of recent papers have utilized transient histograms from commodity SPADs to perform scene inference. Each of these works uses sensors which are very similar to the one used in this paper in terms of technology, form factor, and cost. Callenberg et al. \cite{Callenberg2021CheapSPAD} propose the use of transient histograms from a SPAD to classify materials based on subsurface scattering (with the sensor placed in direct contact), generate higher resolution depth imagery, and perform non-line-of-sight imaging (with additional hardware). Becker and Koerner \cite{becker2023plastic} also classify materials, but in a non-contact setting. Ruget et al.  \cite{ruget2022pixels2pose} perform super resolution and use supervised machine learning to estimate human poses from transient histograms. Other works also perform super resolution to resolve higher resolution depth images from relatively few transient histograms \cite{bian2022large, ruget2021real}.

The differentiable rendering approach used in this work is inspired by Jungerman et al. \cite{jungerman2022}, who use differentiable rendering to recover partial plane parameters from a single transient histogram. Because the sensor used by Jungerman et al. reports only a single transient histogram, only two of the three planar degrees of freedom could be recovered from a single sensor measurement. In contrast, the multiple transient histograms reported by the sensor used in this work enable recovery of all plane parameters from a single measurement, and our work is the first to do so.



\section{Forward Imaging Model}
\label{sec:forward_model}
An accurate forward imaging model is crucial to enabling our differentiable rendering method. In this section, we give an overview of our forward imaging model, which is designed for the TMF8820 sensor, but can in principle be adapted to other sensors. Our model assumes planar scene geometry, with uniform albedo and reflectance model parameters per-plane. For a more general forward imaging model that is sensor agnostic, refer to previous work \cite{jungerman2022}. We consider a set of transient histograms which are simultaneously captured by a transient sensor over different fields-of-view. We refer to this set of histograms as an \textit{image} $\Phi$. Each image consists of $n$ histograms $\varphi \in \Phi$. Each histogram consists of $m$ bins, $\varphi_i : 1 \leq i \leq m$.

\subsection{Surface Reflection Model}
We utilize the Phong reflection model \cite{phong1975illumination}, in which a surface's reflection properties are parameterized by its albedo $\alpha$, specular exponent $k_e$, and specular weight $k_s$. We assume that the light source and sensor are co-located, the pulsed laser source is the only light in the scene, and the strength of illumination is uniform over the field of view. The intensity $I$ of incident light returned by a ray $\vb{r} \in \mathbb{R}^3$ intersecting with plane $\vb{a}\vb{x} + d = 0$ is given by:
\begin{equation}
    I = \alpha * (1-k_s)(\vb{r} \cdot \vb{a}) + k_s  ((2(\vb{r} \cdot \vb{a})\vb{a} + \vb{r})\cdot \vb{r})^{k_e}
    \label{eqn:intensity}
\end{equation}


\subsection{SPAD Saturation}
The Phong reflection model alone does not take into account light falloff or unique properties of the SPAD sensor. Previous work \cite{gupta2019photon} has established that, due to the nature of SPADs, the number of detected photons $p$ follows a soft saturation curve in relation to the number of incident photons $\phi$, given by $p = 1-e^{-\phi}$. Due to the inverse-square law, a ray which travels distance $r$ from the sensor before bouncing off the scene returns with an intensity of $1/r^2$. We incorporate the plane's albedo $\alpha$, as well as the output $I$ from the lighting model given in \cref{eqn:intensity}. The asymptotic highest possible photon detection count $\sigma$ is a property of the sensor. The sensor gain parameter $g$ scales the intensity for an individual ray--this is included because in practice we do not simulate as many rays as photons are actually measured by the sensor. The number of detected photons $p$ is then given by:
\begin{equation}
    p = \sigma(1-e^{-g I/(\sigma r^2)})
    \label{eqn:full_falloff}
\end{equation}

\subsection{Histogram Formation}
Consider a histogram $\varphi$ which images a plane given by $\vb{a}\vb{x} + d = 0$, with a uniform albedo and reflectance parameters. Let the sensor reside at the origin, and let $R$ be a set of rays uniformly sampled from the field-of-view which $\varphi$ images. If $\varphi$ has $n$ bins, a bin temporal ``size'' of $t$, and a bin offset $\omega$ (meaning a flight time of t is recorded as $t + \omega$), the value of an individual histogram bin is given by:
\begin{equation*}
        i_{s} = \omega + t(i-1) \quad \quad i_{e} = \omega+ti
\end{equation*}
\begin{equation}
    \varphi_i^{raw} = \sum_{\vb{r} \in R} \begin{cases}
        p(\vb{r}) & \text{if} \ i_{s} \leq \dfrac{2 || \texttt{isect}(\vb{r}, \vb{a}, d) ||_2}{c} < i_{e}\\
        0 & \text{otherwise}
    \end{cases}
\end{equation}
Where $p(\vb{r})$ is the intensity of light returned by ray $\vb{r}$, as given in \cref{eqn:full_falloff}, $c$ is the speed of light, and $\texttt{isect}(\cdot) \in \mathbb{R}^3$ is the intersection point of $\vb{r}$ with $\vb{a}\vb{x} + d = 0$. Because the sensor that we model (TMF8820) filters out ambient light on-sensor, we assume no ambient light in our imaging model.

\subsection{Laser Impulse}
\begin{figure}
    \centering
    \includegraphics[width=\linewidth]{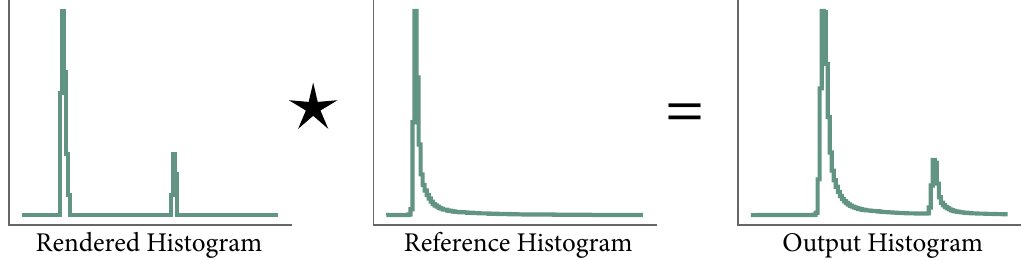}
    \caption{The raw rendered histogram is cross-correlated with the reference histogram (which encodes the laser pulse intensity over time) to generate the output histogram of our forward imaging model.}
    \label{fig:impulse}
    \vspace{-1.5em}
\end{figure}

The sensor which we model records the intensity of its laser impulse over time by piping the laser pulse directly to a SPAD \cite{TMF8820}, and reports the result as a ``reference histogram''. The captured transient histogram is effectively temporally blurred by a kernel matching the reference histogram. To replicate this effect, we cross-correlate the reference histogram with the generated histogram as a step in our forward process, as shown in \cref{fig:impulse}. In the case of the TMF8820 sensor that we utilize, the temporal scale (bin size) is not the same in the reference histogram $\delta$ as in the transient histogram $\varphi$, so we temporally scale $\delta$ by a factor $s_\delta$ before applying the cross-correlation. The histogram after correlation is given by
\begin{equation}
    \varphi^{corr} = \varphi^{raw} \star \texttt{rescale}(\delta, s_{\delta})
\end{equation}
Where $\star$ denotes the cross-correlation operation, and the \texttt{rescale} function scales the function $\delta$ temporally by $s_{\delta}$.

\subsection{Inter-histogram Interference}
The sensor that we model suffers from \textit{inter-histogram interference}, meaning light detected in one histogram is also detected in other histograms, scaled by a factor. We assume that one histogram interferes with all other histograms with an equal magnitude $\psi$, meaning that a bin value of $x$ in one histogram will manifest as $\psi x$ in all other histograms. Formally, for a histogram $\varphi \in \Phi$, where $\varphi_i$ is the $i\textsuperscript{th}$ bin of $\varphi$,
\begin{equation}
    \varphi_i = \varphi_i^{corr} + \psi  \sum_{\tilde{\varphi}^{corr} \in \Phi^{corr}} \tilde{\varphi}_i^{corr}
\end{equation}

\section{Differentiable Rendering Pipeline}
\label{sec:differentiable_rendering}
Our differentiable rendering pipeline recovers plane parameters by minimizing the loss between an observed histogram and the output of a render function. The render function renders a histogram image $\Phi^r$ as a function of four sets of variables: the scene geometry $G$, reflectance model parameters $F$, the sensor's forward imaging model parameters $C$, and the sensor's reported laser impulse $\delta$:
\begin{equation}
    \Phi^r = R(G, F, C, \delta)
\end{equation}

The render function assumes that the sensor is placed at the origin and the optical axis is aligned with the positive z axis. The planar geometry $G$ is given by the angle of incidence $\theta$ of the optical axis to the plane, the intersection point of the plane with the z axis $Z_0$, and the azimuth angle $\phi$, which denotes rotation about the optical axis.

The lighting parameters $F$ are comprised of the Phong reflection model parameters ($k_s, k_e, \alpha$). The camera parameters $C$ are comprised of those described in \cref{sec:forward_model} ($n, t, g, \psi, s_{\delta}, \sigma$), along with 36 scalar parameters which define the height, width, and center of each of the sensor's 9 FoV regions. These FoV parameters are derived from the TMF8820 specification sheet \cite{TMF8820}, and are not differentiable in our implementation. Also included in $C$ is an integer which denotes the number of random ray samples used to render the transient histogram. We keep this fixed at 2304 per FoV region. The impulse response function $\delta$ is reported by the sensor along with every image.

To compare the rendered histogram $\Phi^r$ to the observed histogram $\Phi^{o}$, the following loss function $\mathcal{L}$ is used:
\begin{equation}
    \mathcal{L}(\Phi^r, \Phi^{o}) = \sum_{(\varphi^r, \varphi^{o}) \in \Phi^r, \Phi^{o}} \left\lvert\left\lvert\dfrac{\varphi^r}{\texttt{max}(\varphi^{o})} - \dfrac{\varphi^{o}}{\texttt{max}(\varphi^{o})}\right\rvert\right\rvert_2
    \label{eqn:loss}
\end{equation}
Dividing by the magnitude of the observed histogram ensures that high magnitude histograms do not dominate the loss. Unlike previous work \cite{jungerman2022}, we do not use a Fourier transform-based loss function. In our tests, the L2-norm function above performed slightly better. We believe this is because we utilize a good initial estimate from the histogram peak based approach described in \cref{sec:hist_peak_approach}. A Fourier-based loss excels when the rendered and observed histograms are very different, but may not provide as strong of a signal when they are similar. We adapt Mu et al.'s Python implementation \cite{mu2022physics} of the algorithm given by Urea et al. \cite{Urena2013} to uniformly sample rays from the rectangular FoV of the TMF8820. 

The process of assigning a value to a histogram bin is inherently non-differentiable, as there is an instantaneous change in the output histogram as the input crosses a bin boundary. Following previous work \cite{jungerman2022}, we make the render function differentiable via a soft binning process, in which a Gaussian kernel is generated centered at each input datapoint, and these Gaussians sampled at the bin centers and summed across the datapoint dimension to generate an approximation of the histogram. In our implementation, each Gaussian is also weighted according to its intensity, which is given by \cref{eqn:full_falloff}. The same soft binning process is used to temporally scale the reference histogram $\delta$.

To tune the parameters of our forward imaging model, we utilize a large dataset $D$ of ($\Phi^o, \delta$) pairs, each with an associated ground truth geometry $G$. We minimize the reconstruction loss, as given in \cref{eqn:loss} over the entire dataset to find the ideal camera parameters $C^*$:
\begin{equation}
    C^* = \underset{C}{\arg\min}\sum_{(\Phi^o, G, \delta) \in D} \mathcal{L}(R(G, F, C, \delta), \Phi^o)
    \label{eqn:opt_forward_params}
\end{equation}
Where the reflectance model parameters $F$ are free variables. This optimization only needs to be performed once, as $C^*$ remains fixed for a given sensor.

To recover the geometric parameters $G^*$ of a planar surface from a single image $\Phi^o$ with reference histogram $\delta$, we use the optimized forward imaging parameters $C^*$, while allowing the scene geometry $G$ and reflectance model parameters $F$ to change:
\begin{equation}
    G^*, F^* = \underset{G, F}{\arg\min} \quad \mathcal{L}(R(G, F, C^*, \delta), \Phi^o)
    \label{eqn:plane_recovery_opt}
\end{equation}

Performing this optimization also recovers the reflectance model parameters $F^*$ of the surface. We evaluate the consistency of the surface albedo recovered by this approach in section \cref{subsec:albedo_recovery_results}. Recovery of other reflectance parameters is left for future work as it is outside of the scope of this paper, and evaluation of these parameters is difficult.

Optimization is performed via stochastic gradient descent using the Adam optimizer \cite{kingma2014adam}. The render function $R$ is implemented in PyTorch with gradients generated through automatic differentiation. To initialize $G$ in \cref{eqn:plane_recovery_opt}, we use the output of the histogram peak based approach described in \cref{sec:hist_peak_approach}. We observe that regardless of what reasonable starting estimate is used, the optimization tends to converge to the same solution for planar geometry.

\section{Histogram Peak Based Approach}
\label{sec:hist_peak_approach}
We provide an empirically calibrated approach which is able to approximate the performance of differentiable rendering on the plane recovery task. This method operates by estimating the distance to the plane in each field of view, projecting outwards by the distance, and fitting a plane to the projected points. To tune the method, we optimize a linear mapping from histogram bin to distance (given by parameters $m$ and $b$ below). We also optimize the angle at which points are projected outwards; a different angle is used depending on whether the histogram corresponds to a field of view region on the edge or corner of the overall $3\times3$ region field-of-view ($s_e$ or $s_c$ respectively). The algorithm for this approach is shown in \cref{alg:ad_hoc}.

\begin{algorithm}
    \begin{algorithmic}
        \Function{RecoverPlane}{$\Phi$, $m$, $b$, $s_e$, $s_c$}
            \State $pts \gets [\ ]$
            \For {$\varphi$ in $\Phi$}
                \State $i \gets$ the temporal coordinate of the peak of $\varphi$
                \State $dist \gets i * m + b$
                \State $\vb{u} \gets $ unit vector pointing to center of FoV of $\varphi$
                \If{$\varphi$ images an edge FoV region}
                    \State Scale angle of $\vb{u}$ from optical axis by $s_e$
                \ElsIf{$\varphi$ images a corner FoV region}
                    \State Scale angle of $\vb{u}$ from optical axis by $s_c$
                \EndIf
                \State $pt \gets \vb{u} * dist$
                \State Append $pt$ to $pts$
            \EndFor
            \State Fit a plane to $pts$ via SVD \cite{brown1976principal}
            \State \textbf{return} the parameters $\vb{a}, d$ of the fit plane
        \EndFunction
    \end{algorithmic}
    \caption{Empirically calibrated algorithm for recovering planar geometry from a set of transient histograms using histogram peaks}
    \label{alg:ad_hoc}
\end{algorithm}

To find the location of the peak in a histogram $\varphi$, we fit a piecewise cubic curve to the 128-bin histogram, and sample that curve at 10$\times$ density around the highest individual bin. The temporal position of the highest point on the interpolated curve is the location of the peak. This process captures variations smaller than the $\sim1.2$cm equivalent bins of the histogram by using the relative intensity between adjacent bins. We find empirically that this approach outperforms picking the highest bin without interpolation.

To determine the optimal parameters $m$, $b$, $s_e$, and $s_c$ to the RecoverPlane function, we find the parameters which minimize the error in the reconstructed plane over some calibration dataset $D$ which contains images $\Phi$ along with ground truth planar geometry $\vb{a}, d$:
\begin{equation}
    m^*, b^*, s_e^*, s_c^* = \underset{m, b, s_e, s_c}{\arg\min} \sum_{\Phi, \vb{a}, d \in D} \epsilon_p(f(\Phi, m, b, s_e, s_c), \vb{a}, d)
    \label{eqn:optimize_peak_approach}
\end{equation}
\noindent
where the $\epsilon_p$ is the point error between two planes, as defined in \cref{eqn:point_error}, and $f$ is the RecoverPlane function given in \cref{alg:ad_hoc}. We perform this optimization using the Nelder-Mead method \cite{Nelder1965ASM} with finite difference estimation of derivatives, via the SciPy Python library. As this optimization is performed only once, speed is not crucial.

\section{Experimental Results}
\subsection{Sensor Configuration}
We run the TMF8820 in ``short range, high accuracy'' mode, in which it reports 128 bins with an individual bin size equivalent to $\sim1.2$cm of distance. We run the sensor in the default configuration of 4 million iterations (light pulses) per measurement, and use the default field-of-view configuration, which gives an FoV of $33\degree\times34\degree$, divided into $3\times3$ regions, with a transient histogram reported for each region.

\subsection{Metrics}
We use three metrics to measure the accuracy of plane recovery. Assume that we are comparing two planes given by $\vb{a}_1\vb{x}+d_1=0$ and $\vb{a}_2\vb{x}+d_2=0$, where $d_1 > 0$, $d_2 > 0$, then the \textit{angular error} $\epsilon_a = \arccos(\vb{a}_1 \cdot \vb{a}_2)$. \textit{Linear error} is given by $\epsilon_l = |d_1 - d_2|$. These metrics are intuitive, but the trade-off between the two is not clear. To capture error with a single metric, we define point error $\epsilon_p$. Given a random ray originating at the sensor and within the sensor's FoV, point error captures the expected difference between the intersection of that ray with the predicted plane and with the ground truth plane. Formally:
\begin{equation}
    \epsilon_p = \dfrac{\sum_{\vb{r} \in R} || \texttt{isect}(\vb{a}_1, d_1, \vb{r}) - \texttt{isect}(\vb{a}_2, d_2, \vb{r}) ||_2}{|R|}
    \label{eqn:point_error}
\end{equation}
where \texttt{isect}$(\vb{a}, d, \vb{r})$ returns the 3D point of intersection between plane $\vb{a}\vb{x} + d = 0$ and ray $\vb{r}$, and $R$ is a randomly sampled set of rays originating at the sensor and within the sensor's FoV. In practice, we set $R$ to be an $8\times8$ grid of rays which uniformly cover the sensor's FoV for repeatability.

\begin{figure}
    \centering
    \includegraphics[width=0.8\linewidth]{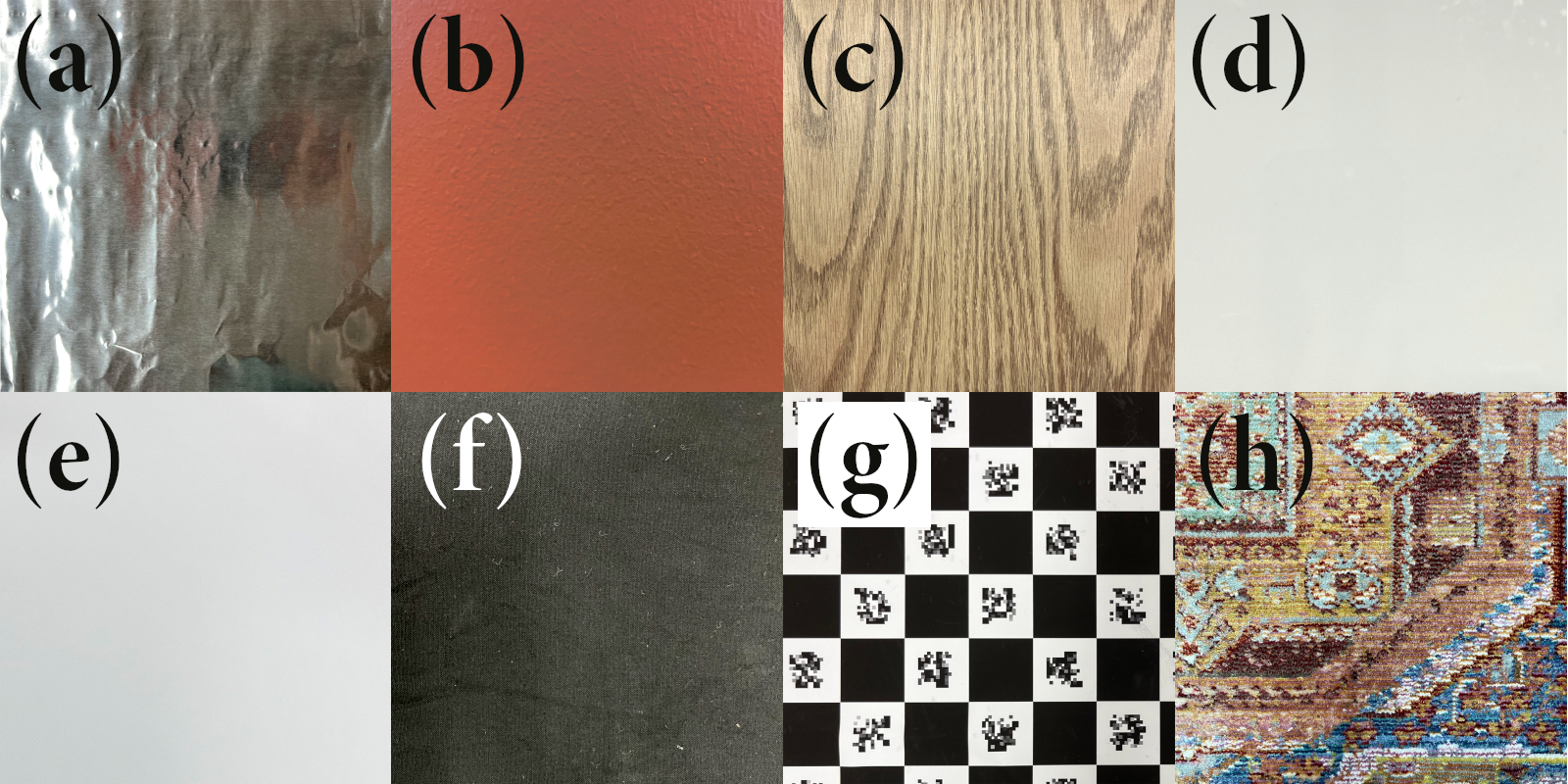}
    \caption{\textbf{Materials on which we evaluate planar recovery.} (a) aluminum foil; (b) red painted drywall; (c) wooden table; (d) whiteboard; (e) white paper; (f) black fabric; (g) checkerboard; (h) patterned rug.}
    \label{fig:materials}
    \vspace{-1.5em}
\end{figure}

\subsection{Planar Recovery}
\label{subsec:expreiments-plane-geometry}
We evaluate five different approaches for planar recovery:
\begin{enumerate}
    \item Differentiable rendering, the optimization problem in \cref{eqn:plane_recovery_opt} is solved.
    \item Peak finding - calibrated, the histogram peak based approach given by \cref{alg:ad_hoc} is performed with optimized parameters given by \cref{eqn:optimize_peak_approach}. 
    \item Proprietary distances - calibrated, the same as 2), but utilizing distance estimates generated onboard the sensor rather than histogram peak locations.
    \item Peak finding - naive, the histogram peak based approach is used, but without optimized parameters.
    \item Proprietary distances - naive, the same as 4), but utilizing distance estimates generated onboard the sensor rather than histogram peaks.
\end{enumerate}

To generate ground truth data, we mount a TMF8820 sensor to a custom 3D printed end effector for a Universal Robots UR5 robot arm. We manually calibrate the position of the sensor relative to the end effector, and use the robot's forward kinematics (which are quoted as precise to $\pm$ 0.1mm) to gather ground truth sensor poses. To determine the position of the plane relative to the robot, the end effector is touched to the plane at a number of points, and a ground truth plane is fit to these points. The robot is used to automatically move the sensor, allowing us to generate a large dataset of planar images (3,800 images total) from a variety of distances ($Z_0$), angles-of-incidence ($\theta$), and azimuth angles ($\phi$). All measurements were captured in an artificially lit room.

To ensure the validity of our results when comparing differentiable rendering to other approaches, we use the worst-performing naive proprietary distances approach as a starting estimate, and perform 100 iterations of gradient descent. One iteration takes about 0.05 seconds on a mid-range laptop (i7-10705H, NVIDIA GTX 1650Ti). In real-world operation, a better starting estimate could be used and fewer iterations performed. The peak-based approaches operate at 95Hz on the same hardware, exceeding the 30Hz at which the sensor reports data.

\begin{table*}
    \centering
    \begin{tabular}{l | R R R | R R R | R R R}
         & \multicolumn{3}{c|}{Angular Error (\degree)} & \multicolumn{3}{c|}{Linear Error (mm)} & \multicolumn{3}{c}{Point Error (mm)} \\
         Method & Mean & Median & 95\% & Mean & Median & 95\% & Mean & Median & 95\% \\
         \hline
         \rule{0pt}{1.1\normalbaselineskip}%
         Differentiable Rendering\textsuperscript{*} & \textbf{3.40} & \textbf{1.97} & \textbf{12.90} & \textbf{2.46} & \textbf{1.90} & \textbf{6.51} & \textbf{3.79} & \textbf{3.17} & 8.46 \\
         Peak Finding - Calibrated\textsuperscript{\dag} & 3.57 & 2.22 & 13.44 & 2.67 & 2.11 & 7.13 & 3.94 & 3.52 &\textbf{7.92} \\
         Peak Finding - Naive & 5.68 & 3.87 & 18.42 & 6.15 & 5.28 & 13.56 & 7.70 & 6.96 & 14.28 \\
         Proprietary Distances - Calibrated\textsuperscript{\dag} & 7.34 & 4.71 & 25.97 & 49.20 & 60.31 & 68.41 & 52.44 & 62.96 & 69.60 \\
         Proprietary Distances - Naive & 8.87 & 4.71 & 30.06 & 60.31 & 71.96 & 78.14 & 65.45 & 76.26 & 83.15 \\  
    \end{tabular}
    \caption{\textbf{Methods which utilize the histogram outperform those which use proprietary distance estimates in all metrics.} Images in range 1-30cm to plane, 0-30\degree AoI on surfaces (c) - (h). 400 measurements per surface. Measurements of surface (b) from the same range were used optimize forward model of differentiable method (*) and calibrate ``calibrated'' methods (\dag). 95\% refers to the 95\textsuperscript{th} percentile of error. See \cref{subsec:expreiments-plane-geometry} for a description of methods.}
    \label{tab:plane_recovery}
\end{table*}

\begin{table}
    \centering
    \begin{tabular}{l|r r r}
         & \multicolumn{3}{c}{Point Error (mm)}\\
         Method & Mean & Median & 95\% \\
         \hline
         \rule{0pt}{1.1\normalbaselineskip}%
         Differentiable Rendering\textsuperscript{*} & \textbf{6.26} & \textbf{3.52} & \textbf{22.31} \\
         Peak Finding - Calibrated\textsuperscript{\dag} & 6.80 & 3.78 & 23.58 \\
         Peak Finding - Naive & 15.84 & 11.45 & 44.56 \\
         Proprietary Distances - Naive & 42.23 & 22.15 & 130.46 \\
         Proprietary Distances - Calibrated\textsuperscript{\dag} & 74.45 & 75.45 & 143.51 \\
    \end{tabular}
    \caption{\textbf{Methods which utilize the histogram outperform those which use proprietary distance estimates in larger range of plane parameters.} Measurements cover range 1-70cm, 0-45\degree AoI of surface (b). Measurements of surface (e) from range 0-30cm, 0-30\degree \ AoI were used optimize forward model of renderer (*) and to calibrate ``calibrated'' methods (\dag).}
    \label{tab:recovery_wider_range}
    \vspace{-1em}
\end{table}

\begin{figure}
    \centering
    \includegraphics[width=\linewidth]{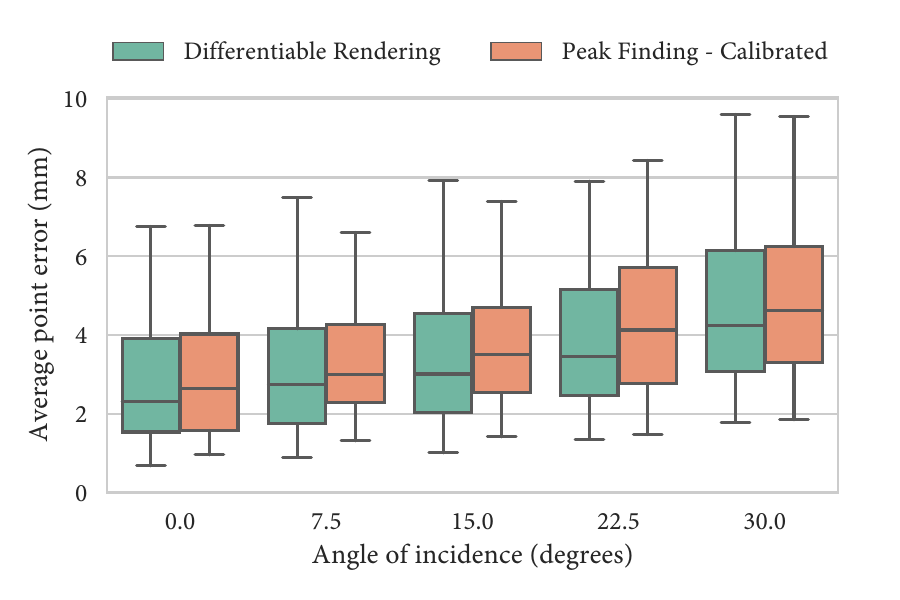}
    \caption{\textbf{Higher angle-of-incidence leads to higher error in reconstruction.} Measurements of materials (c)-(h) cover distance range 0-30cm. Whiskers extend to 5\textsuperscript{th} and 95\textsuperscript{th} percentile.}
    \label{fig:aoi_vs_error}
    \vspace{-1em}
\end{figure}

\begin{figure}
    \centering
    \includegraphics[width=\linewidth]{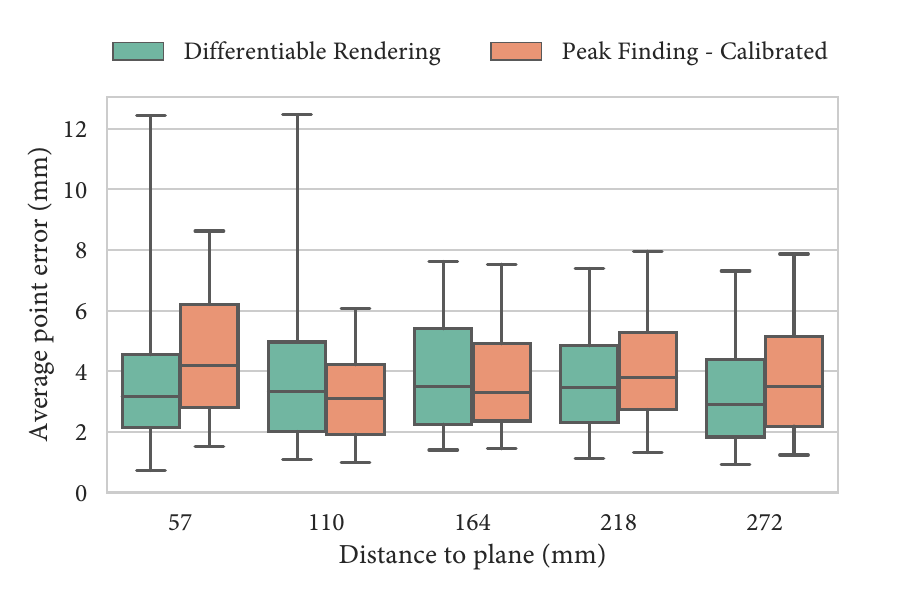}
    \caption{\textbf{Distance to the planar surface has little effect on reconstruction error.} Measurements of materials (c)-(h) cover AoI range 0-30\degree. Whiskers extend to 5\textsuperscript{th} and 95\textsuperscript{th} percentile. Ticks on x axis denote center of 54mm bins.}
    \label{fig:dist_vs_error}
    \vspace{-1.5em}
\end{figure}


A comparison between the five approaches is given in \cref{tab:plane_recovery}. Methods which utilize transient histograms consistently outperform those which rely on proprietary distance estimates. We see that the differentiable rendering approach, which utilizes the entirety of the information in all nine histograms, outperforms peak finding approaches, in which each histogram is reduced to a single value. Our peak finding approach comes close to the performance of differentiable rendering across the board, even outperforming it in some cases, offering speed at the expense of generality. We believe the large gap between the ``peak finding'' and ``proprietary distances'' approaches can partially be explained by a difference in interpolation method; the interpolation method used onboard the sensor may be less accurate than the one used in our peak finding method. However, in our testing we found that even when using \textit{no interpolation at all}, our peak finding approach outperformed the proprietary distances approaches, necessitating an additional explanation.

We suspect that the proprietary algorithm onboard the sensor is not overly naive, but instead is designed to be more general purpose than our approach. Plane fitting is a special case; an algorithm which performs well for a variety of potential use cases may not be optimal for plane fitting. The peak finding method that we use was chosen \textit{because} it is effective at recovering planar surfaces. By accessing the transient histograms directly, we were afforded the ability to make this choice. Results of planar recovery over a wider range of sensor poses are shown in \cref{tab:recovery_wider_range}. The effect of angle-of-incidence (AoI) and distance on reconstruction error is shown in \cref{fig:aoi_vs_error} and \cref{fig:dist_vs_error}.

\begin{table}
    \centering
    \begin{tabular}{p{8em} | G G G}
         & \multicolumn{3}{c}{Mean Point Error (mm)}\\
         \hspace{8em} Material & Diff. Render & Peak Find & Propr. Dist. \\
         \hline
         \rule{0pt}{1.1\normalbaselineskip}%
         (b) Red drywall* & \textbf{2.05} & 3.09 & 4.68\\
         (e) White paper & \textbf{2.51} & 3.45 & 63.0\\
         (f) Black fabric & \textbf{2.51} & 3.35 & 72.5\\
         (h) Patterned rug & \textbf{2.69} & 3.62 & 62.7\\
         (c) Wood & 4.19 & \textbf{4.03} & 60.7\\
         (d) Whiteboard & \textbf{5.12} & 5.82 & 54.9\\
         (g) Checkerboard & 6.94 & \textbf{4.12} & 61.1\\
         (a) Aluminum foil & \textbf{12.7} & 15.0 & 25.3\\
    \end{tabular}
    \caption{\textbf{Our methods are generally robust to surface properties, aside from highly specular aluminum foil.} Images in range 1-30cm, 0-30\degree \ AoI. *measurements of red drywall are used to optimize forward model of differentiable method and to calibrate peak finding and proprietary distance approaches.}
    \label{tab:multiple_surfaces}
    \vspace{-1em}
\end{table}

We evaluate our method on a range of surfaces and report the results in \cref{tab:multiple_surfaces}. The parameters of the ``calibrated'' methods and imaging model parameters of the differentiable rendering methods were trained on measurements of the red painted drywall. We see that our methods are generally robust to this change from training to testing surface, particularly when that surface has a uniform texture and albedo. Our methods are slightly less robust to textured surfaces such as wood and a patterned rug. We see diminished performance with the slightly glossy whiteboard, and the checkerboard surface, which has spatially varying albedo. Performance is significantly diminished on the specular aluminum foil.

We observe that the proprietary distance based approach tends to overfit when calibrated to a dataset. There is evidence of this overfitting in the longer range test in \cref{tab:recovery_wider_range}; the calibrated histogram approach improves over the naive approach, while the calibrated proprietary distance approach performs \textit{worse} than the naive. This is because the parameters of the ``calibrated'' approaches were calibrated to recover planar geometry on images of a different surface over a different range of distances and angles of incidence. While the histogram based approaches, including differentiable rendering, are robust to this change in surface, the approaches which utilize proprietary distances are not.

\subsection{Albedo Recovery}
\label{subsec:albedo_recovery_results}
\begin{figure}
    \centering
    \includegraphics[width=\linewidth]{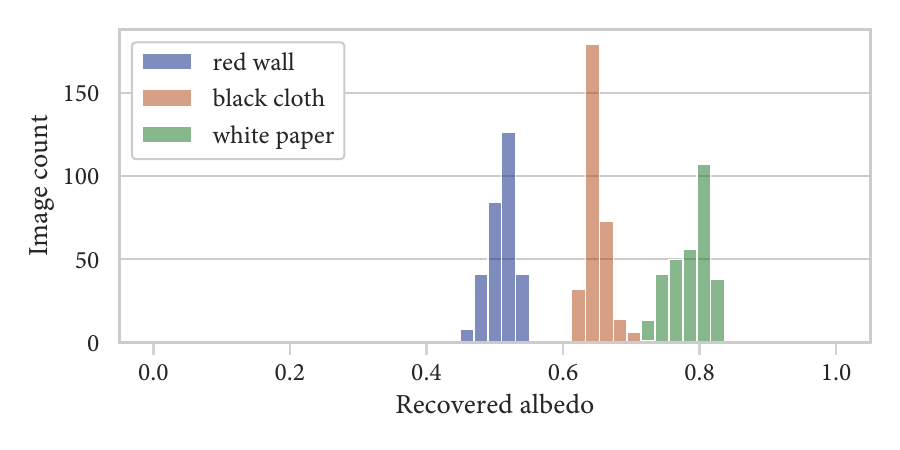}
    \caption{\textbf{Our method recovers consistent surface albedo under various distances and angles-of-incidence.} Recovered albedo is in the wavelength of the sensor light source (940nm IR), and may vary significantly from the albedo as it appears to the human eye under visible light. Each surface is observed from 300 poses in range 7-40cm, 0-30\degree \ AoI.}
    \label{fig:albedo_hist}
    \vspace{-1em}
\end{figure}

We evaluate the performance of our differentiable renderer for recovering surface albedo, as given in \cref{eqn:plane_recovery_opt} by recovering the albedo from images of three planar surfaces which have a uniform texture and albedo. We only evaluate the \textit{consistency} of the recovered albedo, not the \textit{accuracy}. Evaluating the accuracy would require an accurate characterization of the wavelength of the sensor light source, and surfaces with a known albedo in that wavelength, which is outside the scope of this work. We find that this method recovers a consistent albedo per-surface relatively invariant to distance in the range 7-40cm and angle-of-incidence in the range 0-30\degree, allowing discrimination between surfaces, as shown in \cref{fig:albedo_hist}.

\section{Example Application}
\label{sec:demonstration}
We build an application to showcase our methods, in which a robot arm is holding a cup of liquid. The robot's goal is to safely place the cup on a tabletop below, which is at an unknown distance and may have regions which are not level. In our application, we attach a TMF8820 transient sensor directly to the gripper of the robot arm. Due to its small size, the sensor can be placed centimeters away from the jaws of the gripper, where it senses the surface below directly, making it invulnerable to occlusions.

\begin{figure}
    \centering
    \includegraphics[width=\linewidth]{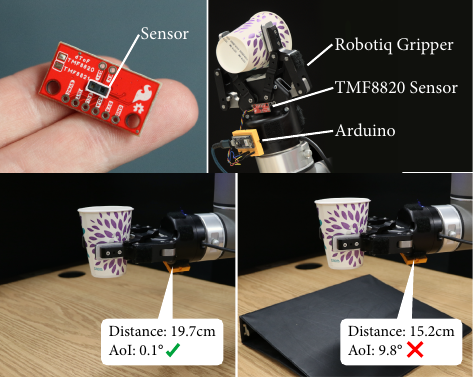}
    \caption{We mount a proximity sensor on a robot gripper (top). The sensor detects when the surface below the gripper is level and safe to place a cup full of liquid (bottom left) or is not level and therefore unsafe (bottom right).}
    \label{fig:demo}
    \vspace{-1.5em}
\end{figure}

Using our approach for recovering planar geometry, the robot is able to sense the distance to and slope of the surface below the cup being held in the end effector, as shown in \cref{fig:demo}. The robot uses this information to know when it is close enough to the surface to place the cup down, and to ensure that the surface is level enough to safely release the cup.

\section{Limitations}
While the differentiable rendering method given in this work can in principle recover any unknown parameters to the render function, we only evaluate recovery of scene geometry and albedo. A next step is to investigate recovery of the reflectance model parameters of a surface. While our method in principle enables such recovery, evaluation is difficult. Another next step is recovering other types of geometry. Our approach can in principle easily be adapted to other parameterized surfaces, \textit{e.g.}, a sphere or cube. Extending to arbitrary geometry would require a more general differentiable representation, \textit{e.g.} a neural representation akin to NeRF \cite{mildenhall2021nerf}. As both of these tasks introduce extra degrees of freedom into the optimization process, they may require a more accurate and/or sophisticated model of the transient histogram imaging process to sufficiently constrain optimization.

One challenge for future work is the low bandwidth available on commodity sensors. In our test setup, histograms are read from the sensor at 4.5 frames per second (where one ``frame'' consists of nine histograms) despite the sensor generating proximity estimates at 150Hz. This is not a limitation of the sensor technology, but of the I\textsuperscript{2}C interface over which it transmits data. We hope that commodity SPAD sensors will in the future come packaged with high bandwidth interfaces to enable granular and high-speed sensing. Algorithms will also need to be optimized to perform inference quickly enough to keep pace with higher bandwidth sensors.

Lastly, we provide only a basic demonstration of utilizing transient histograms in a robotics setting. It is yet to be shown that utilizing these histograms leads to improvement in performance of downstream robotics tasks. An important next step is to build a complete robotics system which utilizes transient histogram data, and evaluate the system performance compared to alternative sensing modalities. We are hopeful that future robotics systems will harness the power of transient histograms to be highly aware of their environment on a low size, weight, and power budget.


\bibliography{references}
\bibliographystyle{IEEEtran}

\end{document}